\documentclass[sigconf]{acmart}

\usepackage{booktabs} 
\usepackage{epstopdf}
\usepackage[utf8]{inputenc}

\usepackage{hyperref}
\usepackage{xstring}

\usepackage{xcolor}

\usepackage{comment}
\definecolor{wikiblue}{HTML}{0645AD}
\definecolor{red}{HTML}{D21919}
\usepackage{graphicx}
\usepackage{tabularx}

\usepackage{pgfplots}
\usepackage[caption=false]{subfig}
\usepgfplotslibrary{external}

\setcopyright{acmcopyright}

\acmDOI{xx.xxx/xxx_x}

\acmISBN{978-1-4503-9517-5/23/03}

\acmConference[SAC'23]{ACM SAC Conference}{March 27 –April 2, 2023}{Tallinn, Estonia}
\acmYear{2023}
\copyrightyear{2023}

\acmArticle{4}
\acmPrice{15.00}

\begin{document}
\title{FREDA: Flexible Relation Extraction Data Annotation}
  
\renewcommand{\shorttitle}{FREDA: Flexible Relation Extraction Data Annotation}

\author{Michael Strobl}
\orcid{0000-0002-4062-1026}
\affiliation{%
  \institution{University of Alberta}
  \city{Edmonton} 
  \country{Canada} 
}
\email{mstrobl@ualberta.ca}

\author{Amine Trabelsi}
\affiliation{%
  \institution{Lakehead University}
  \city{Thunder Bay} 
  \country{Canada} 
}
\email{atrabels@lakeheadu.ca}

\author{Osmar Za\"iane}
\affiliation{%
  \institution{University of Alberta}
  \city{Edmonton} 
  \country{Canada} 
}
\email{zaiane@ualberta.ca}

\renewcommand{\shortauthors}{Strobl et al.}

\begin{abstract}
To effectively train accurate Relation Extraction models, sufficient  and  properly labeled data is required. Adequately labeled data is difficult to obtain and annotating such data is a tricky undertaking. Previous works have shown that either accuracy has to be sacrificed or the task is extremely time-consuming, if done accurately. We are proposing an approach in order to produce high-quality datasets for the task of Relation Extraction quickly. Neural models, trained to do Relation Extraction on the created datasets, achieve very good results and generalize well to other datasets. In our study, we were able to annotate 10,022 sentences for 19 relations in a reasonable amount of time, and trained a commonly used baseline model for each relation.
\end{abstract}

%
%



\maketitle

\section{Introduction}
\label{sec:introduction}

The task of Relation Extraction (RE) is one of the major parts of Knowledge Base Population (KBP) \cite{tac2011}, i.e. the augmentation of an existing Knowledge Base (KB). The main goal is to recognize relations, which are expressed between two entities mentioned in the same sentence or document. At present, this is usually achieved by a model based on neural networks, which is trained on, ideally, large amounts of labeled data. However, there are very few publicly available labeled datasets. When available, these are usually limited to specific relations, and often lack the relations that one is interested in. 
This leads to the question on how new datasets for RE can be created. Typically this is a tradeoff between time efficiency and accuracy. Through crowd-sourcing, large manually annotated datasets, such as the TAC Relation Extraction Dataset (TACRED) \cite{zhang2017position}, can be created relatively quickly, although quality may be questionable. Recently, \citet{alt-etal-2020-tacred} gave some insights on TACRED, suggesting that more than 50\% of the challenging examples, i.e. trained models make mistakes on, may need to be relabeled. On the other hand, carefully manually annotated datasets with multiple annotators, such as KnowledgeNet \cite{mesquita2019knowledgenet}, are extremely time-consuming to create and therefore often not feasible. In this work, we propose an approach that makes it possible to create high-quality datasets with a moderate amount of time and effort.

To illustrate the difficulty of the annotation task for RE, consider the following example\footnote{From Melinda Gates' Wikipedia article \url{https://en.wikipedia.org/wiki/Melinda_Gates}}:

\begin{quote}
    ``\textcolor{red}{\textbf{Melinda}} began dating \textcolor{olive}{\textbf{Microsoft}} CEO \textcolor{wikiblue}{\textbf{Bill Gates}} in 1987, after meeting \textcolor{wikiblue}{\textbf{him}} at a trade fair in \textcolor{magenta}{\textbf{New York}}.''
\end{quote}

Four entities are mentioned (\textcolor{red}{\textbf{Melinda}}, \textcolor{olive}{\textbf{Microsoft}}, \textcolor{wikiblue}{\textbf{Bill Gates}} with his co-reference \textcolor{wikiblue}{\textbf{him}}, and finally \textcolor{magenta}{\textbf{New York}}). If the task is detecting the \textit{ceo\_of}-relation, the fact is explicit in the sentence. However, if the task is detecting the \textit{spouse}-relation, one may indicate such relationship between Melinda and Bill Gates. While this might be true, it is only based on the annotator's knowledge and such annotation could improperly mislead a classifier and  train it to perhaps associate ``dating'' with the \textit{spouse}-relation.    

Another possible annotation error can be illustrated by this example:
\begin{quote}
``\textcolor{red}{\textbf{Bill Gates}} has received an honorary Doctorate from \textcolor{wikiblue}{\textbf{Cambridge}}.''
\end{quote}

An annotator confused about the meaning of {\bf alma mater} could erroneously tag the \textit{alma\_mater}-relation between Gates and the university while Bill Gates never attended Cambridge. Although not always done (see TACRED \cite{zhang2017position}), this problem could be mitigated by using multiple annotators, as suggested by \citet{alt-etal-2020-tacred}.

This leads to the question, how is it possible to create RE datasets, which:
\textbf{(1)} are of high enough quality and validated;
\textbf{(2)} are large enough to train effective models;
\textbf{(3)} can be relatively easily extended with more relations;
\textbf{(4)} can be quick to annotate and construct;

We believe these four conditions are essential for the creation of useful RE datasets.

Previous work encountered some difficulties to fulfill these conditions.
\citet{alt-etal-2020-tacred} suggest that generating high quality and validated data (condition (1)) may not be met through crowd-sourcing or at least not if there are no measures in place that ensure data quality. Therefore, ideally, multiple annotators per example are necessary to ensure consistency. In addition, \citet{rosenman2020exposing} suggested that all entities in a sentence need to be annotated in order to train models that generalize properly, which is not always done. 

Creating large enough datasets (condition (2)) is not a trivial task and the required size of labeled datasets is usually unknown in advance. Indeed, labeling entities, co-references, and their relations to each other, even in a single sentence, can be complicated, depending on the relation in question, as well as time-consuming \cite{mesquita2019knowledgenet}. Moreover, existing datasets often contain a predetermined rigid set of common annotated relations like \textit{located\_in}, \textit{founded} and \textit{spouse}. Therefore, an easily extendable and flexible framework (condition (3)) is practical to gather data for new specific or uncommon relations. Enabling a quick construction (condition (4)) is not about the haste in annotation, which in turn could lead to errors, but more about the ease of annotation to avoid a repetitious tedious task. This ease of annotation is conducive to the collection of larger datasets (condition (2)). 

Therefore, there is the need for an easy-to-use framework for sentence annotation for RE, with which the aforementioned conditions for such datasets can be met. These are the main contributions of this paper:

\begin{itemize}
    \item We propose a framework, FREDA for Flexible Relation Extraction Data Annotation, which can be used to manually annotate sentences quickly and accurately. A simple procedure for sentence acquisition from a partially annotated Wikipedia-based corpus is provided to be able to create datasets for new relations.
    \item We provide a dataset with 10,022 sentences annotated for 19 relations (15 used by \citet{mesquita2019knowledgenet} and 4 new relations) with at least two annotators per sentence (third annotator for tie-breaking in case the first two disagree). Models trained on these sentences and their labels show a significant performance gain in F1 scores than previously reported results on common RE datasets, demonstrating that it is possible to obtain significantly better results when the annotations are of high quality. 
\end{itemize}

The remainder of this article is structured as follows: Section \ref{sec:related} presents related work on RE dataset annotation. Section \ref{sec:annotation} describes our annotation approach. Section \ref{sec:models} delineates the model architecture. Section \ref{sec:evaluation} details the evaluation procedure with our conclusions in Section \ref{sec:conclusion}.

\section{Related Work}
\label{sec:related}

The TAC Relation Extraction Dataset (TACRED) \cite{zhang2017position} is a large dataset which used Mechanical Turk crowd annotation with 41 relations and 106,264 examples. However, each example was only annotated by a single annotator and, as pointed out by \citet{alt-etal-2020-tacred}, there is a large number of labeling errors misleading trained models. Although TACRED is still a popular dataset and widely used, e.g. by \citet{joshi2020spanbert} or \citet{alt2018improving}, presumably since previous semi-automatic labelling approaches, such as text annotations with Distant Supervision (see \citet{riedel2010modeling}), i.e. aligning sentences with facts from KBs only through matching entities, are even more error-prone. It seems to be important to double check annotations with more annotators, even though less text can be annotated that way.

In addition, \citet{rosenman2020exposing} investigated the heuristics that a model trained on TACRED may learn to score high on the test set without solving the underlying problem: 
(1) Only 17.2\% of the sentences in TACRED have more than a single pair of entities annotated. Therefore, in most sentences a model will only encounter a single pair of entities, for which it is asked to predict a relation. Instead of predicting a relationship for this pair, it may rather learn to predict whether a sentence expresses a certain relationship, ignoring the potential subject and object. This would lead to a high recall, but may also result in many false positives, leading to a low precision when tested.

(2) A classifier may predict a relation solely based on whether the types of entities in that relation are present in a sentence, especially for relations, which have a unique entity-type-pair, such as \textit{per:religion} with PERSON as subject and RELIGION as object. The authors did a manual investigation of this relation with models trained on TACRED, often leading to false positives, if entities of type PERSON and RELIGION were present in sentences from Wikipedia, but the relation actually not expressed. One of their conclusions was that all entities in a sentence need to be annotated to lower the impact of these problems on the trained model, i.e. a model may generalize better to unseen data in this case.


KnowledgeNet \cite{mesquita2019knowledgenet} is a project aiming to manually annotate 100,000 facts for 100 properties, although at the time of publication 13,425 facts for 15 relations were available. Data annotation consists of the following steps:

\begin{enumerate}
    \item Fetch sentences: Using T-REx \cite{elsahar2018t}, a system to align text and KB facts, to find sentences that could describe facts from Wikidata\footnote{\url{https://www.wikidata.org}} and, in addition, sentences that contain certain keywords.
    \item Mention Detection: Annotators are asked to highlight entity names.
    \item Fact Classification: Pairs of mentions are classified as positive or negative for a relation.
    \item Entity Linking: Linking mentioned entities to their corresponding Wikidata entity.
\end{enumerate}

Each sentence is labeled by at least two annotators to ensure high data quality. The authors report an average of 3.9 minutes to annotate a single sentence by up to 3 annotators. In our approach, Entity Linking is not explicitly done. However, this step is only responsible for 28\% of the time spent according to their study, and therefore a significant amount of time (about 2.8 minutes) is still needed to annotate a sentence. 

There is also a variety of web-based annotation tools available, open-source tools, such as BRAT \cite{stenetorp2012brat}, as well as proprietary ones, such as Prodigy\footnote{\url{https://prodi.gy/}}, which can be used for RE data annotation, among a variety of other NLP tasks. However, these tools are complicated to set up and use, making RE data annotation a time-consuming task, similar to \cite{mesquita2019knowledgenet}.

\section{Data Annotation with FREDA}
\label{sec:annotation}

Some works, such as \citet{yu-etal-2020-dialogue}, claim that most facts span multiple sentences since only Named Entities (NE) can be considered as subjects or objects. However, entities are usually referenced by pronouns or other co-references in the same sentence that expresses a property. Therefore, we claim that a sentence-based approach, like ours, 
is adequate in this case. The same one-sentence-based approach has also been applied by \citet{mesquita2019knowledgenet}, even though a following Entity Linking step was used.

In general, manually creating  labeled datasets for RE involves three tasks: (1) data acquisition, (2) data filtering (sentences in our case) and (3) the actual annotation task. Our approach for each of these tasks is described below.

\subsection{Sentences from WEXEA}

A major part of RE data annotation is selecting entities and their co-references, which could potentially be involved in a relation. Therefore, a corpus should be used with entities already labeled. This would only leave the decision on which entities are subject/object and whether a relation is expressed to the annotator. For example, a corpus for Named Entity Recognition (NER), such as the CONLL 2003 dataset \cite{sang2003introduction}, could be used. However, manually labeled corpora are usually limited in size and it is unlikely that they contain enough sentences that could be relevant for a specific set of relations.

To avoid this issue of potentially running out of text data, we are using sentences from WEXEA (Wikipedia EXhaustive Entity Annotations)~\cite{strobl2020wexea}. WEXEA is an exhaustively annotated dataset derived from the English Wikipedia, which currently contains over 6,000,000 articles. Wikipedia already includes many hyperlinks to entities. However, some may still be missing. Indeed, Wikipedia editors are not encouraged to link the same entity twice, the entity an article is about, and widely known entities, whose link is obvious to the reader. 
However, WEXEA authors claim to capture all these missing links since it is easier than in, for example, news text, which typically does not contain any links to start with. Therefore, WEXEA can be used in our approach since it contains entity annotations already, which can speed up the process. In addition, articles are already split into sentences. 

WEXEA does not contain dates and times, but since some relations, e.g. \textit{date\_of\_birth}, require these, we use the SUTime library \cite{chang2012sutime}, implemented within the CoreNLP tool \cite{manning-EtAl:2014:P14-5}, in order to add these as entities.

\subsection{Sentence Filtering}
\label{ssec:filtering}

If all available sentences are considered for annotation for each relation, the percentage of relevant sentences\footnote{It is somewhat subjective what a relevant sentence is for a specific relation. It could be a sentence which contains entities of both entity types participating in the relation or a sentence being somehow related to the relation topic-wise.} is expected to be very low.
This is an issue since the number of relevant sentences for each relation is presumably very imbalanced for a corpus without pre-selection, as it is the case for TACRED, for example. The by far most common relation in this dataset is \textit{per:title} with 3,862 examples, whereas 37 out of 41 relations have less than 1,000 examples with 4 relations having even less than 100. Sentence filtering ensures that each relation has enough relevant examples to make sure a model can be trained on all relations properly.
Therefore, we are using a similar approach as used by \citet{mesquita2019knowledgenet}:

\begin{itemize}
    \item Keywords: We define a set of keywords relevant to each relation, which are used to filter sentences. This is done for each relation separately. We chose relevant words for each relation including WordNet synonyms \cite{miller1995wordnet}. The keywords used by \citet{mesquita2019knowledgenet} were not reported.
    \item Distant Supervision: Since it is difficult to define an exhaustive set of keywords, a Distant Supervision approach is used to find sentences not matching these keywords, but still containing entities, which are known to be related to each other with a relation of interest, e.g. the sentence mentioning Bill and Melinda Gates as seen in Section \ref{sec:introduction}. 
    Distant Supervision is typically used to add positive examples before training. In our case, we use distance supervision to add candidate sentences for annotation. If for a pair of entities a knowledge base indicates the existence of a relationship, we use these two entities as a query to find sentences as candidates for annotation for the stated relation. We are using DBpedia \cite{10.5555/1785162.1785216} as KB and extracted all entity pairs for each relation, which can be found in DBpedia. 
\end{itemize}

Distant Supervision for a specific relation can only be used if this relation is part of a DBpedia, which is not the case for all relations we annotated sentences for. But if it was the case, up to 50\% of the relevant sentences are extracted this way.

\subsection{Sentence Annotation Task}

\begin{figure*}[t!]
\includegraphics[width=\textwidth]{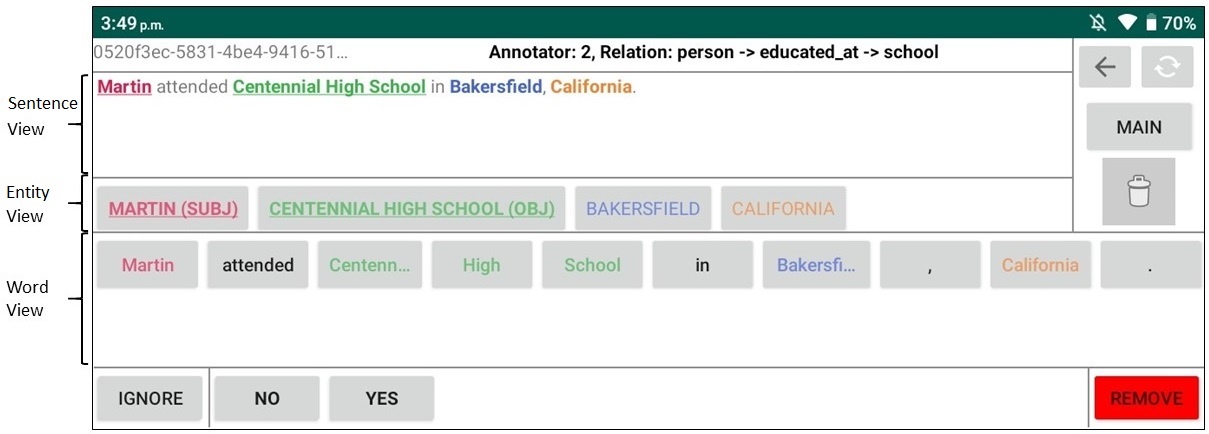}
\caption{Interface for RE data annotation FREDA. Relation considered here is ``educated at'' with ``Martin'' as subject and ``Centennial High School'' as object. The locations ``Bakersfield'' and ``California'' are not participating in the relation, but can be used for creating negative examples.}
\label{fig:app}
\end{figure*}

Figure \ref{fig:app} provides an overview of the mobile annotation application of FREDA.
The objective of the tool is to facilitate the annotation task and reduce the cognitive load for its users. This would lead to collecting annotations of decent quantities and high quality in a relatively short amount of time.

Given a particular relation, the tool initially provides a sentence with entity annotations highlighted in different colours in the \textit{Sentence View}. These initial entity annotations are extracted and loaded by leveraging the WEXEA dataset and the SUTime library. The \textit{Entity View} contains distinct entity buttons which are unique labels representing each entity annotated in the \textit{Sentence View}. Thus, if multiple different mentions or co-references to the same entity occur in a sentence (\textit{Sentence View}), they would all be represented by one entity button in the \textit{Entity View}. The colour of that entity button and all corresponding mentions in the \textit{Sentence View} would be the same. This would help reduce ambiguity and facilitate the decision making. Another practical advantage of the tool consists of reducing the indication of the subject (SUBJ) and the object (OBJ) entities in the relation to just a simple press of the corresponding entity buttons in the \textit{Entity View}. In other words, it is not necessary to look through every single mention of an entity and indicate the role. Annotating only the representative entity button in the \textit{Entity View} is sufficient.  

The \textit{Word View} contains buttons representing every single token in the sentence at hand. By using this view, conducting different editing operations becomes straightforward. For instance, new entities can be easily created through dragging and dropping one of the word buttons in the \textit{Word View} up into the \textit{Entity View}. Similarly, entities can be removed or fixed (e.g. adding a missing word) by just dragging and dropping. Web-based tools, e.g. BRAT\cite{stenetorp2012brat}, require the user to select spans of text using a computer mouse to create entities, requiring very precise (i.e. slow) moves.

In most sentences, editing operations using FREDA require a very minimal amount of time. Once done with editing, a user indicates whether the relation holds or not, i.e. a simple binary decision is made at the end. Sentences are considered for multiple relations as long as they meet the criteria outlined in Section \ref{ssec:filtering}.  An annotator can also remove the sentence from the database (e.g. sentence is broken or list items) or ignore for the current relation.

\subsection{Multiple Annotators}

For such a complex task it is expected that a single annotator is not able to be accurate and consistent over hundreds of sentences. \citet{alt-etal-2020-tacred} show a detailed analysis of all the mistakes that are possible, especially for crowd-sourced annotation tasks, where presumably time matters more than accuracy. Therefore, similar to \citet{mesquita2019knowledgenet}, our system relies on at least two annotators per sentence with a third annotator to break ties if the previous two disagree in their decision. The second annotator gets to see the entity annotations from the first annotator (and the third from the second). Note that only the entity annotations are carried over in this way. Thus, the final decision as well as indicating the subject and object entities is still every annotator's independent decision. But carrying over entity annotations saves time for subsequent annotators (also removed/ignored sentences are not shown to annotators thereafter). Therefore, the time spent for subsequent annotators is likely to be lower than for the first. In this work, the annotators were mainly researchers in the field of NLP.

\section{Model Architectures}
\label{sec:models}

\begin{figure}
\centering
 \includegraphics[width=0.47\textwidth]{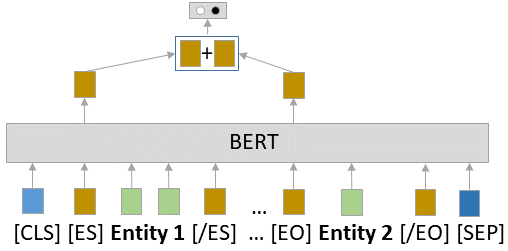}
 \caption{Model used for Relation Extraction from a sentence with annotated subject and object.}
 \label{fig:model}
\end{figure}

We consider the best model architecture from \citet{soares2019matching}, which is based on the BERT Transformer model \cite{devlin2019bert} and showed good results on TACRED. Although instead of a multi-class classifier, as commonly used for models trained on TACRED, we altered the model to do binary classification. It is depicted in Figure \ref{fig:model}. Apart from BERT's special tokens, two new tokens are introduced for the subject (entity start token \textit{[ES]} and entity end token \textit{[/ES]}) and for the object (entity start token \textit{[EO]} and entity end token \textit{[/EO]}), which are referred to as entity markers. The BERT embeddings of the start tokens of both entities concatenated are used as input of a classification layer with a sigmoid activation function, which makes a binary decision whether the relation of interest is expressed between the marked entities. Since sentences can express multiple relations, even between the same entity-pair, independent classification decisions have to be made and one model per relation can be trained. Therefore, the input of the model for a given relation is a sentence with one entity mention marked as subject (with \textit{[ES]} and \textit{[/ES]}) and one entity mention marked as object (with \textit{[EO]} and \textit{[/EO]}). 

This model architecture aims to solely recognize and decide whether or not the context of two entities suggests that the relation, the model was trained for, is expressed. There are no additional tasks such as NER or co-reference resolution that have to be learned and may influence the quality of the predictions. Therefore, this architecture should be suitable for finding out what performance is possible for the task of RE if accurately annotated and large enough datasets  are available for training.


\section{Evaluation}
\label{sec:evaluation}

In this Section, we provide a detailed evaluation of how models trained on datasets created with FREDA perform, how much data is required to reach a certain performance level for selected relations and how fast these annotations can be acquired.

\subsection{Model Training and Dataset Statistics}

We are using the previously mentioned model architecture outlined in Figure \ref{fig:model} with the cased large model of BERT. Learning rate of $5*10^{-6}$ (linear decay), Adam optimizer \cite{kingma2014adam}, batch size 32, 1-10 epochs (varies per relation; determined using 5-fold cross-validation). Test sets contain 10\% of the whole dataset, one per relation.

Table \ref{tab:statistics} shows statistics of our datasets. 7 annotators annotated in total 10,022 sentences with $>$500 for 19 relation. 15 of these relations can be found in the KnowledgeNet dataset as well. In addition, we annotated 4 more common relations, which are also part of the schema.org ontology for persons\footnote{\url{https://schema.org/Person}}. \textit{Positive responses} refers to the number of sentences, which were deemed as ``expressing the relation'', whereas \textit{negative responses} correspond to sentences ``not expressing the relation'' for any pair of entities. From these positive responses, 
a number of \textit{positive facts} can be extracted with a positive label, which can be used for model training. Often it is possible to extract multiple such facts for a single sentence since subjects and objects can be mentioned several times in the sentence and we may have different subjects or objects in the same sentence. 

For instance, consider the following sentence:

\begin{quote}
``\textcolor{red}{\textbf{Princess Alberta}} was the fourth daughter of \textcolor{wikiblue}{\textbf{Queen Victoria}} and \textcolor{olive}{\textbf{Prince Albert}}.''
\end{quote}

Two positive facts can be extracted for the \textit{child\_of} relation: 
\begin{itemize}
    \item \textcolor{red}{\textbf{Princess Alberta}} \textit{child\_of} \textcolor{wikiblue}{\textbf{Queen Victoria}}
    \item \textcolor{red}{\textbf{Princess Alberta}} \textit{child\_of} \textcolor{olive}{\textbf{Prince Albert}}
\end{itemize}

Facts with a negative label can be easily created by considering all other pairs of entities, which do not express the relation at hand. Therefore, four negative facts can be extracted from the previous sentence for the same relation\footnote{It is possible that these negative facts still express another relation, e.g. \textit{parent} or \textit{spouse}. But since the corresponding model is trained to do binary classification, they are considered as negative facts in this case.}: 
\begin{itemize}
    \item \textcolor{wikiblue}{\textbf{Queen Victoria}} \textit{child\_of} \textcolor{red}{\textbf{Princess Alberta}}
    \item \textcolor{olive}{\textbf{Prince Albert}} \textit{child\_of} \textcolor{red}{\textbf{Princess Alberta}}
    \item \textcolor{olive}{\textbf{Prince Albert}} \textit{child\_of} \textcolor{wikiblue}{\textbf{Queen Victoria}}
    \item \textcolor{wikiblue}{\textbf{Queen Victoria}} \textit{child\_of} \textcolor{olive}{\textbf{Prince Albert}}
\end{itemize}

Since there are typically more negative facts than positive ones, each training example is weighted in the loss function (binary cross-entropy loss) accordingly, in order to account for the class imbalance.

We calculated the inter-annotator agreement for the first and second annotator with Cohen's Kappa.
Two annotators agree when they both consider that the relation in question is expressed (or inexistent) in the sentence at hand. Overall, the results can be considered as excellent with $\kappa = 0.85$ for all relations together. Although it ranges between 0.48 (\textit{place\_of\_residence} and 0.96 (\textit{date\_of\_birth}). Some relations require more discussion between annotators are therefore more difficult and time-consuming to annotate for humans, leading to more disagreement.

\begin{table}[ht!]
\centering
\begin{tabular}{|l|l|}
\hline
\textbf{Statistics} & \textbf{Total}\\ \hline
Relations & 19 \\
Sentences & 10,022\\
Positive \mbox{responses} & 5,371\\
Negative \mbox{responses} & 4,651\\
\mbox{Positive} \mbox{facts} & 11,160 \\
\mbox{Negative} \mbox{facts} & 232,678\\
Inter-annotator kappa & 0.85 \\
\hline
\end{tabular}
\caption{Data statistics (Total): Number of sentences, number of yes- and no-responses, number of positive and negative facts extracted from these and inter-annotator kappa (between the first two annotators).}
\label{tab:statistics}
\end{table}

\subsection{Test Results}

We trained models on all datasets created with FREDA and tested them on the corresponding test sets as well as on unseen data provided by KnowledgeNet, which can be considered as high-quality.

The KnowledgeNet training data can be downloaded from their repository\footnote{\url{https://github.com/diffbot/knowledge-net}}, which can be used for testing models trained on our datasets. Since this dataset does not contain exhaustive entity annotations (only entities participating in a specific relation are annotated), negative examples for testing can only be generated from sentences expressing a relation. These are presumably the more challenging sentences since the model needs to figure out which entity is subject, which is object and which entities are neither. Also, entity-pairs for negative examples can be extracted through considering mentions of the same entity, which cannot be related to each other. For all relations, this results in 10,895 positive and 46,347 negative examples, compared to 11,160 positive and 232,678 negative examples for our datasets\footnote{Each sentence can contain multiple positive and negative examples, each subject-object-combination is considered.}.

We are reporting Precision, Recall and F1 score on the FREDA and KnowledgeNet test sets in Table \ref{tab:results}, broken up for each relation as well as \textbf{Interim} results for relations which can be found in the datasets from both approaches, and the \textbf{Total} which includes the four additional relations we annotated with FREDA.

The \textbf{Interim} F1 score 
of 0.86
on the FREDA test sets is 
relatively high overall. Even though it is not possible to directly compare this result against results of state-of-the-art models trained on the commonly used TACRED dataset, these latter usually achieve a significantly lower F1 score on TACRED\footnote{\url{https://paperswithcode.com/sota/relation-extraction-on-tacred}}. Models trained on these datasets created with FREDA also show a similarly high F1 score on the KnowledgeNet dataset with 0.87. Although these datasets were created by different annotators and presumably similar, but still, in detail, different approaches for sentence filtering.

It is often not reported, but we can gain some insights on how such models perform on different relations. While relations, which do not leave a lot of room for interpretation, such as \textit{date\_of\_birth}, \textit{place\_of\_birth}, \textit{child\_of}, \textit{spouse} or \textit{sibling}, show a very high F1 score, others, such as \textit{subsidiary\_of} or \textit{place\_of\_residence}, show significantly worse results. Cohen's Kappa for inter-annotator agreement for these two relations is quite low with 0.64 and 0.48, respectively, compared to the average of 0.85. Therefore, the datasets corresponding to these two relations can be considered as more challenging to train on, and thus more sentences may be needed.

\begin{table*}[ht!]
\centering
\begin{tabular}{|l|lll|lll|}
\hline
\textbf{}                & \multicolumn{6}{c|}{\textbf{Datasets}}  \\ \hline
\textbf{}                & \multicolumn{3}{c|}{\textbf{FREDA (test)}}    & \multicolumn{3}{c|}{\textbf{KnowledgeNet}} \\ \hline
\textbf{Relation}        & \textbf{P} & \textbf{R} & \textbf{F1} & \textbf{P}   & \textbf{R}  & \textbf{F1}  \\ \hline
date\_of\_birth (PER \textrightarrow DATE)         & 0.96 & 0.97 & 0.96 & 0.90 & 1.00 & 0.94         \\
date\_of\_death (PER \textrightarrow DATE)         & 0.93 & 0.96 & 0.94 & 0.93 & 0.92 & 0.93             \\
place\_of\_residence (PER \textrightarrow LOC)     & 0.71 & 0.76 & 0.73 & 0.86 & 0.74 & 0.79              \\
place\_of\_birth (PER \textrightarrow LOC)         & 0.85 & 1.00 & 0.92 & 0.95 & 0.81 & 0.87              \\
nationality (PER \textrightarrow LOC)              & 0.84 & 0.95 & 0.89 & 0.92 & 0.92 & 0.92             \\
employee\_or\_member\_of (PER \textrightarrow ORG) & 0.68 & 0.91 & 0.78 & 0.95 & 0.82 & 0.88             \\
educated\_at (PER \textrightarrow ORG)             & 0.87 & 0.94 & 0.90 & 0.98 & 0.90 & 0.94             \\
political\_affiliation (PER \textrightarrow ORG)   & 0.96 & 1.00 & 0.98 & 0.90 & 0.90 & 0.90             \\
child\_of (PER \textrightarrow PER)                & 0.75 & 0.83 & 0.79 & 0.91 & 0.89 & 0.90             \\
spouse (PER $\leftrightarrow$ PER)                 & 0.93 & 0.91 & 0.92 & 0.95 & 0.89 & 0.92             \\
date\_founded (PER \textrightarrow DATE)           & 0.83 & 0.95 & 0.89 & 0.94 & 0.88 & 0.91             \\
headquarters (ORG \textrightarrow LOC)             & 0.80 & 0.84 & 0.82 & 0.94 & 0.86 & 0.90             \\
subsidiary\_of (ORG \textrightarrow ORG)           & 0.51 & 0.71 & 0.59 & 0.87 & 0.74 & 0.80              \\
founded (PER \textrightarrow ORG)                  & 0.72 & 0.94 & 0.82 & 0.49 & 0.82 & 0.61              \\
ceo\_of (PER \textrightarrow ORG)                  & 0.81 & 0.89 & 0.85 & 0.94 & 0.91 & 0.93            \\ \hline
\textbf{Interim}                                   & 0.83 & 0.90 & 0.86 & 0.88 & 0.86 & 0.87 \\ \hline
award (PER \textrightarrow AWARD)                  & 0.78 & 0.83 & 0.80 & --   &  --  & --              \\
alma\_mater (PER \textrightarrow ORG)              & 0.70 & 0.62 & 0.65 & --   &  --  & --              \\
place\_of\_death (PER \textrightarrow LOC)         & 0.79 & 0.90 & 0.84 & --   &  --  & --              \\
sibling (PER $\leftrightarrow$ PER)                & 0.77 & 0.79 & 0.78 & --   &  --  & --              \\ \hline
\textbf{Total}                                     & 0.82 & 0.89 & 0.85 & --   &  --  & --             \\
\hline
\end{tabular}
\caption{Test set results of the models trained on the FREDA training sets for each relation and both approaches. The last 4 relations are not part of KnowledgeNet's dataset, therefore the results are missing. \textbf{Interim} corresponds to the overall results for all 15 relations in both datasets. \textbf{Total} includes results on all relations in FREDA's test set.}
\label{tab:results}
\end{table*}

\subsection{Challenge RE dataset}

\begin{table*}[ht!]
\centering
\begin{tabular}{|l|lll|lll|}
\hline
\textbf{}                & \multicolumn{6}{c|}{\textbf{Models}}  \\ \hline
\textbf{}                & \multicolumn{3}{c|}{\textbf{FREDA}}    & \multicolumn{3}{c|}{\textbf{KnowBERT-W+W}} \\ \hline
\textbf{Relation}        & \textbf{P} & \textbf{R} & \textbf{F1} & \textbf{P}   & \textbf{R}  & \textbf{F1}  \\ \hline
date\_of\_birth    & 0.96 & 0.93 & \textbf{0.95} & 0.67 & 0.99 & 0.80         \\
date\_of\_death    & 0.74 & 0.78 & \textbf{0.76} & 0.61 & 0.74 & 0.67         \\
educated\_at       & 0.85 & 0.72 & 0.78          & 0.68 & 0.93 & \textbf{0.79}         \\
sibling            & 0.76 & 0.87 & \textbf{0.81} & 0.53 & 0.89 & 0.67         \\
spouse             & 0.84 & 0.87 & \textbf{0.85} & 0.56 & 0.86 & 0.68         \\
founded            & 0.86 & 0.53 & 0.66          & 0.82 & 0.76 & \textbf{0.79}  \\
date\_founded      & 0.86 & 0.60 & 0.71          & 0.60 & 0.89 & \textbf{0.72} \\ \hline
\textbf{Total}     & 0.83 & 0.76 & \textbf{0.79} & 0.63 & 0.87 & 0.73          \\
\hline
\end{tabular}
\caption{Challenge RE dataset test results. The previously trained models from FREDA were used as well as the KnowBERT-W+W model, trained on TACRED and showing state-of-the-art performance on the TACRED test set. The best F1 scores per relation and overall are in bold.}
\label{tab:cre_results}
\end{table*}


\citet{rosenman2020exposing} created a more challenging dataset based on 30 out of 41 TACRED relations, called Challenge RE (CRE). It is relatively balanced, i.e. the number of positive examples is similar to the number of negative examples, whereas TACRED is highly imbalanced. Furthermore, each annotated sentence contains at least two entity pairs that are compatible with the relation the sentence is annotated for, aiming to reveal models that learned to classify sentences rather than classifying entity pairs in a sentence, which would lead to high recall but low precision. Therefore, this dataset is considered to be more challenging than the TACRED test set. CRE was specifically created as a challenging dataset in order to test the generalization capabilities of models, for example, trained on TACRED.

We identified 7 relations in CRE that are fully compatible with 7 of FREDA's relations.
, therefore the previously trained models on FREDA datasets can be used to be tested on the CRE dataset for these relations. In order to compare, we chose the KnowBERT-W+W model from \citet{peters2019knowledge}, a knowledge-enhanced version of BERT through the integration of WordNet \cite{miller1995wordnet} and a subset of Wikipedia. It also uses entity markers for relation prediction and is trained on TACRED and shows state-of-the-art results on the TACRED test set. 

Table \ref{tab:cre_results} shows the results on the CRE dataset for both approaches. Overall, the F1 scores show that models based on FREDA often perform significantly better than KnowBERT-W+W, resulting in a higher total average. Another observation is that KnowBERT-W+W shows a very high recall compared to precision, which is expected due to the nature of TACRED and the resulting lack of generalization when tested on a more challenging dataset, such as CRE. FREDA's models, on the other hand, show a more balanced precision and recall, indicating that they pay more attention to which entity is subject and which is object, i.e. our datasets may lead to better generalization properties for models when trained on them, compared to TACRED. This also indicates that our sentence pre-selection step with keywords and distant supervision, which is important to end up with balanced datasets, is relatively general and does not necessarily only pre-select easy sentences, while still leading to a high F1 score when trained and tested on (see Table \ref{tab:results}).

\subsection{How many sentences do we need per relation?}

TACRED contains a variety of relations with a high variance in the number of examples per relation (3,862 for \textit{per:title} and only 33 for \textit{org:dissolved}). However, typically only the overall performance of a model trained and tested on TACRED is reported in the literature, i.e. it is unknown how well these models perform on each relation and how many examples or sentences per relation are needed to reach a certain performance level.

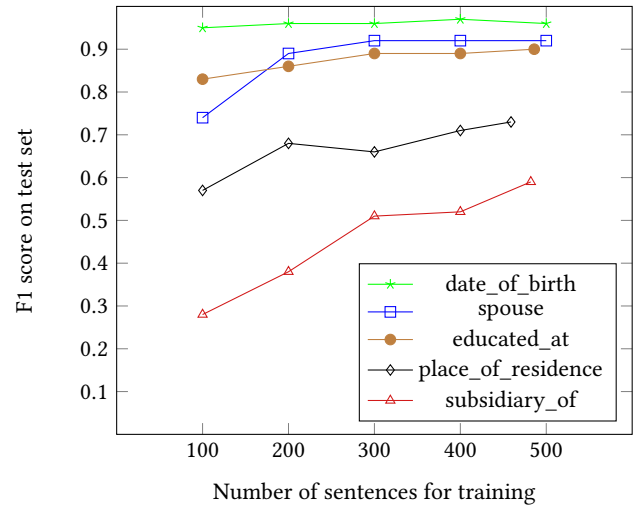
\begin{figure}
\centering
        \begin{tikzpicture}
        \begin{axis}[
            xlabel={Number of sentences for training},
            ylabel={F1 score on test set},
            xmin=0, xmax=600,
            ymin=0, ymax=1.0,
            xtick={100, 200, 300, 400, 500},
            ytick={0.1, 0.2, 0.3, 0.4, 0.5, 0.6, 0.7, 0.8, 0.9, 1.1},
            legend pos=south east,
        ]
        \addplot[
            color=green,
            mark=star,
            ]
            coordinates {
            (100,0.95)(200,0.96)(300,0.96)(400,0.97)(500,0.96)
            };
            \addlegendentry{date\_of\_birth}
        \addplot[
            color=blue,
            mark=square,
            ]
            coordinates {
            (100,0.74)(200,0.89)(300,0.92)(400,0.92)(500,0.92)
            };
            \addlegendentry{spouse}
        \addplot[
            color=brown,
            mark=*,
            ]
            coordinates {
            (100,0.83)(200,0.86)(300,0.89)(400,0.89)(486,0.90)
            };
            \addlegendentry{educated\_at}
        \addplot[
            color=black,
            mark=diamond,
            ]
            coordinates {
            (100,0.57)(200,0.68)(300,0.66)(400,0.71)(459,0.73)
            };
            \addlegendentry{place\_of\_residence}
        \addplot[
            color=red,
            mark=triangle,
            ]
            coordinates {
            (100,0.28)(200,0.38)(300,0.51)(400,0.52)(482,0.59)
            };
            \addlegendentry{subsidiary\_of}
        \end{axis}
    \end{tikzpicture}%
\caption{F1-score on test set for different training set sizes and a selection of relations.}
\label{fig:size}
\end{figure}

We want to shed some light into the question of how many sentences have to be annotated per relation and how is it possible to find out whether more annotated sentences may be beneficial. Figure \ref{fig:size} shows the model performance on the FREDA test sets (same as used for the experiments reported in Table \ref{tab:results}) for five different relations (\textit{date\_of\_birth}, \textit{spouse}, \textit{educated\_at}, \textit{place\_of\_residence} and \textit{subsidiary\_of}), when trained on 100, 200, 300, 400 or all available sentences we annotated using FREDA and shuffled before sampling. These relations were selected since the models trained on the corresponding datasets show different performance levels (Table \ref{tab:results}) as well as the inter-annotator agreement varies widely (Table \ref{tab:statistics}).

The model corresponding to the least controversial relation among annotators (\textit{date\_of\_birth}), i.e. the one with the highest Kappa, already shows stable performance after being trained on only 100 sentences. Models for the \textit{spouse} and \textit{educated\_at} relations need slightly more sentences, but barely improve after being trained on more than 300. Whereas for the \textit{place\_of\_residence} and \textit{subsidiary\_of} relations, even close to 500 sentences\footnote{As previously mentioned, we annotated at least 500 sentences per relation. However, 10\% of these sentences are kept aside as test sets, i.e. the rest of the training sets may contain slightly less than 500 sentences.} seem to be 
insufficient to possibly get to a similarly high performance than the models for the other relations. 

The Pearson Correlation Coefficient between model performance for each relation and the inter-annotator agreement is 0.75, i.e. both values are highly correlated. It can be concluded that if annotators often do not agree on annotations for certain relations, models have more difficulties to predict these relations as well, indicating that more data is needed. Whereas, if annotators barely ever disagree 
a relatively small amount of data is necessary.

\subsection{Annotation Speed}

Data annotation for RE can be prohibitively time-consuming. Therefore, approaches for quick dataset construction are essential in order to be able to easily extend existing datasets with more relations or create entirely new datasets. We asked two annotators, familiar with the task, to annotate sentences for the \textit{spouse}-relation using the following approaches:

\begin{itemize}
    \item \textbf{BRAT} \cite{stenetorp2012brat}: BRAT is an open-source web-based tool for data annotation. Entities of different types can be annotated through selecting spans of text. Relations are annotated through connecting these entities.
    \item \textbf{FREDA (plain)}: In order to solely and fairly compare FREDA's annotation interface with BRAT, WEXEA entity annotations were removed for this approach.
    \item \textbf{FREDA}: Full framework, including annotations from WEXEA.
\end{itemize}

Sentences were randomly selected from WEXEA. In order to keep the workload as similar as possible for each annotator and approach, all sentences contain exactly 25 words and a new unseen set of 100 sentences was used every time. In addition, the annotators were asked to only select entities of relevant types (\textit{Person} in this case), which reduces the workload and the \textit{spouse}-relation is not supposed to be applied to other types. However, WEXEA itself contains entity annotations of other types in which case annotators were not asked to remove entities for the approach \textbf{FREDA}.


\begin{table}[]
\begin{tabular}{|l|l|l|l|l|l|l|}
\hline
         & \multicolumn{2}{c|}{\textbf{BRAT}}  & \multicolumn{2}{c|}{\textbf{FREDA (plain)}} & \multicolumn{2}{c|}{\textbf{FREDA}}  \\ \hline
            & \textbf{sec.} & \textbf{F1} & \textbf{sec.} & \textbf{F1} & \textbf{sec.}  & \textbf{F1} \\ \hline
Annotator A & 23.3 & 0.53 & 17.7 & 0.55  & \textbf{9.2} & \textbf{0.69}  \\ \hline
Annotator B & 33.1 & 0.48 & 25.3 & 0.43 & \textbf{12.4} & \textbf{0.56} \\ \hline
\end{tabular}
\caption{Average annotation speed in seconds per sentence for each annotator, lower is better. A model was trained for each dataset and the F1 score on the CRE dataset for the \textit{spouse}-relation as test set is reported. Best results per annotator are in bold.}
\label{tab:speed}
\end{table}

Table \ref{tab:speed} shows the average annotation speeds in seconds per sentence for both annotators and all three approaches. In general, annotation speeds vary significantly for each annotator since multiple steps are required and the speeds of each of them depend on individual abilities. The resulting datasets were used to train models, which were tested on the CRE dataset for the \textit{spouse}-relation. Annotations with \textbf{FREDA} show the best F1 score for both annotators. Note that entities of other types are annotated as well using \textbf{FREDA}, while still keeping the average time to annotate a sentence low. This resulted in more examples for training and therefore better models when tested on CRE. Results for datasets from the other approaches are similar, suggesting annotation quality is similar\footnote{All F1 scores in Table \ref{tab:speed} are lower than reported in Table \ref{tab:cre_results} since significantly less data was used for training.}.

For both annotators, the FREDA interface, represented through the approach \textbf{FREDA (plain)}, lead to a 24\% increase in annotation speed compared to \textbf{BRAT} and another 48\% to 51\% increase for pre-annotated sentences from WEXEA, i.e. for the approach \textbf{FREDA} compared to \textbf{FREDA (plain)}, even though more entity types were annotated with \textbf{FREDA}. This should give an order of magnitude for the time required to annotated a sentence for a relation and how FREDA can help reduce the workload for annotators through its easy-to-use interface as well as using pre-annotated sentences.

\section{Conclusion}
\label{sec:conclusion}

Previous works on data annotation indicated either that large amounts of data are necessary in order to achieve moderate model performance (see \mbox{TACRED} \cite{zhang2017position}) or data annotation, if done carefully, is extremely time-consuming (see KnowledgeNet \cite{mesquita2019knowledgenet} or BRAT \cite{stenetorp2012brat}). We showed that it is possible to create high-quality datasets for RE for a variety of relations, with a moderate amount of time and effort, using freely available text data from Wikipedia. The resulting models trained on these datasets showed state-of-the-art results for RE and are robust when tested on datasets from different annotators than they were trained on.

We hope that releasing FREDA to the public\footnote{Annotated data, trained models and server and Android application code are publicly available: \url{https://github.com/mjstrobl/FREDA}} will encourage the community to quickly create more annotated data for more relations, which would boost research for the tasks of RE and Knowledge Graph Population.

\bibliographystyle{ACM-Reference-Format}
\bibliography{sample-bibliography} 

\end{document}